\documentclass{esannV2}
\usepackage[latin1]{inputenc}
\usepackage{booktabs}
\usepackage{amssymb,amsmath,array}

\usepackage{multicol}
\usepackage{multirow}
\usepackage{graphicx}
\usepackage{color}
\usepackage{url}

\begin{document}
%style file for ESANN manuscripts
\title{Replay-free Online Continual Learning with Self-Supervised MultiPatches}

%***********************************************************************
% AUTHORS INFORMATION AREA
%***********************************************************************
\author{Giacomo Cignoni$^1$, Andrea Cossu$^1$, Alex G\'omez Villa $^2$, \\Joost van de Weijer$^2$ and Antonio Carta $^1$
%
% Optional short acknowledgment: remove next line if non-needed
\thanks{This paper has been partially supported by the CoEvolution project, funded by EU Horizon 2020 under GA n 101168559 and the EU-EIC EMERGE (Grant No. 101070918).
We acknowledge ISCRA for awarding this project access to the LEONARDO supercomputer, owned by the EuroHPC Joint Undertaking, hosted by CINECA (Italy).}
%
% DO NOT MODIFY THE FOLLOWING '\vspace' ARGUMENT
\vspace{.3cm}\\
%
% Addresses and institutions (remove "1- " in case of a single institution)
1- University of Pisa \ %- Computer Science Department 
%
% Remove the next three lines in case of a single institution
%\vspace{.1cm}\\
2- Computer Vision Center (CVC) \\
}
%***********************************************************************
% END OF AUTHORS INFORMATION AREA
%***********************************************************************

\maketitle

\begin{abstract}
Online Continual Learning (OCL) methods train a model on a non-stationary data stream where only a few examples are available at a time, often leveraging replay strategies. However, usage of replay is sometimes forbidden, especially in applications with strict privacy regulations. Therefore, we propose Continual MultiPatches (CMP), an effective plug-in for existing OCL self-supervised learning strategies that avoids the use of replay samples. CMP generates multiple patches from a single example and projects them into a shared feature space, where patches coming from the same example are pushed together without collapsing into a single point. 
CMP surpasses replay and other SSL-based strategies on OCL streams, challenging the role of replay as a go-to solution for self-supervised OCL.  Code available at \url{https://github.com/giacomo-cgn/cmp} .
\end{abstract}

\section{Introduction}
% \begin{itemize}
%     \item replay is a popular strategy for OCL
%     \item SSL needs large minibatch size and large computational budget
%     \item How to extend each incoming streaming minibatch?
%     \item Replay is a straightforward way to do this, but it suffers from the limitations of memory, especially with small memory buffer
%     \item instead, adapting ssl methods to use multipatch losses is a robust way to achieve this.
% \end{itemize}

The goal of Continual Learning (CL) is the continuous adaptation of deep neural networks to non-stationary streams while accumulating knowledge \cite{rebuffi2017icarl}. In this paper we focus on three desiderata, which are often lacking in state-of-the-art methods: fast adaptation in an online stream, learning without explicit supervision and without access to replay.\\
Recently, there has been a growing interest in Online Continual Learning (OCL) \cite{soutifcormerais2023comprehensive}, a challenging scenario in which the model sees the data in a single pass. At each timestep, the model has access only to a small minibatch of data. As a result, OCL naturally limits the computational budget and requires models that are able to converge quickly with minimal data.
% CL with unlabeled data
Until now, most research in OCL has focused on supervised methods. However, this assumption may be unrealistic since for many real-world applications labels are not immediately available. 
Self-Supervised Learning (SSL) has emerged as an effective paradigm for training deep neural networks from unlabeled data. Previous work in CL even suggests that SSL methods may be more robust to catastrophic forgetting compared to equivalent supervised methods~\cite{cossu2022pretraining, fini2022cassle}.\\
The main limitation of SSL methods is their high computational cost, leading to methods that are slow to converge and require large minibatch sizes.
% main idea
In this paper, we explore Online Continual Self-Supervised Learning (OCSSL), the problem of adapting SSL methods to an OCL scenario~\cite{mai2021onlinesurv}. In this setting, the main challenge arises from the limited amount of data available at each timestep, which results in an overall small computational budget. This issue is usually tackled in OCSSL by leveraging Experience Replay (ER) \cite{chaudhry2019er_online}, which concatenates minibatches from the stream with samples from the memory buffer $\mathcal{M}$.
%The absence of the classification layer in OCSSL reduces the benefits induced by rehearsal, leaving the opportunity to explore alternative ways to increase the minibatch beyond buffers.
%This is also relevant because the introduction of buffers introduces limitations: mainly overfitting on the small subset of samples stored in memory.
% our proposal
Recent work \cite{chen2022bagssl, tong2023emp} introduced the concept of extracting multiple patches from a single image in SSL, with the aim of speeding up the training process. Based on these findings, we introduce Continual MultiPatch (CMP) to extend \textit{Instance Discrimination} SSL methods \cite{gui2023sslsurvey} to multiple patches instead of the standard two views. Like ER, CMP extends the minibatch size by adding multiple patches. However, unlike replay, CMP does not require an external memory buffer and does not store previous data (with advantages for scalability and data privacy). Our experiments on challenging class-incremental OCL benchmarks show that CMP is able to surpass the performance of replay-based strategies as well as comparable OCSSL approaches under a restricted computational budget.

\begin{figure}[t]
    \centering
    \includegraphics[width=0.95\linewidth]{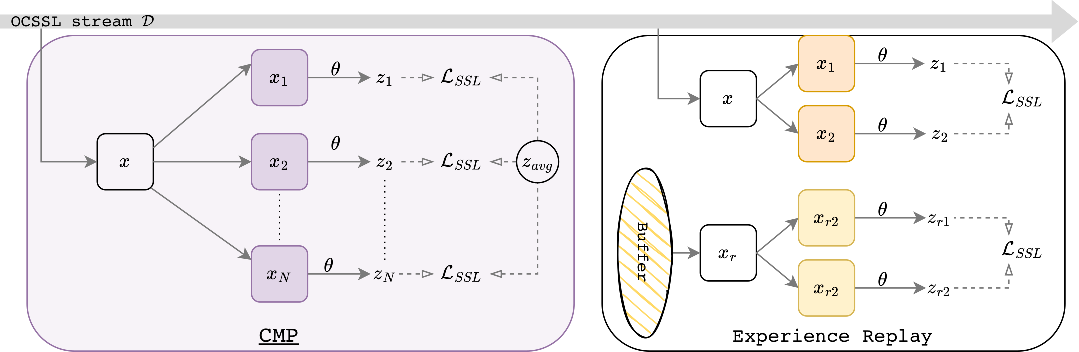}
    \caption{Comparison between CMP (left) and ER (right) in OCSSL. While ER requires an external memory buffer, CMP only requires the current example $x$. 
    % More, CMP loss does not need distinguishing between examples from the buffer and from the stream.
    }
    \label{fig:cmp}
\end{figure}

\section{Related Works}
SSL trains a feature extractor $\theta: \mathcal{X} \rightarrow \mathcal{F}$ to map inputs \(x \in \mathcal{X}\) to latent representations \(z \in \mathcal{F}\). Training involves pretext tasks on unlabeled data, while the evaluation is usually conducted with linear probing on downstream tasks.
We focus on \textit{instance discrimination} SSL methods, where the pretext task aligns two augmented views of the same sample in feature space via contrastive loss \cite{chen2020simclr}, additional predictor head \cite{chen2020simsiam}, clustering \cite{caron2021swav} or redundancy reduction \cite{bardes2022vicreg}.\\
In OCL \cite{soutifcormerais2023comprehensive}, the model faces a non-stationary sequence of data $\mathcal{D} = (\mathcal{D}_1, \mathcal{D}_2, \ldots)$ where each $\mathcal{D}_i$ is composed by a very small number of examples (e.g., usually from $1$ to around $10$). We consider class-incremental data streams \cite{rebuffi2017icarl}, where drifts between a given $\mathcal{D}_i$ and $\mathcal{D}_{i+1}$ introduce examples sampled from unseen classes. Interestingly, in OCL drifts do not occur after each $\mathcal{D}_i$ and the model does not know \emph{when} the drift occurs (boundary-free stream). This contrasts with many SSL methods for CL that require to know in advance when a drift is introduced \cite{gomezvilla2022pfr}. In addition, OCL approaches (both with and without SSL) usually employ replay to increase the amount of examples available at each training iteration and to mitigate forgetting \cite{purushwalkam2022minred, soutifcormerais2023comprehensive, yu2023scale, purushwalkam2022minred}.\\
Our approach is replay-free and works without access to boundaries by leveraging the idea of building multiple patches from a single example. This idea is already present in BagSSL \cite{chen2022bagssl} and EMP-SSL \cite{tong2023emp} but it has not been applied to CL, yet.
EMP-SSL loss enforces similarity between each patch latent representation and their average. EMP-SSL also uses the Total Coding Rate $\mathcal{L}_\textit{TCR}$ (Section \ref{sec:method}) to avoid the collapse of latent representations into a single point.

% \JW{Shall we add a paragraph on existing continual learning methods for self-supervised learning CASSLE etc? Concluding that these are not online.}

\section{Continual MultiPatches} \label{sec:method}
We propose Continual MultiPatches, an SSL method that is i) replay-free, ii) does not require knowledge about boundaries in the stream and iii) is generally applicable on top of Instance Discrimination SSL strategies \cite{gui2023sslsurvey}.
As shown in Figure \ref{fig:cmp}, CMP extracts a set of $N$ patches $x_1, ..., x_N$ by applying different transformations to the original sample $x$.
Then, given an encoder network $\theta$, CMP computes the latent representations $z_1, ...z_N$ for each patch: $z_i = \theta(x_i)$.
Let us call $\mathcal{L}_\textit{SSL}$ the loss of the underlying instance discrimination SSL method, then CMP loss reads:

\begin{equation}
\label{eq:general}
\mathcal{L}_\textit{CMP} = \beta\mathcal{L}_{TCR}([z_1, ..., z_N]) + \alpha\sum_{i=1}^N \mathcal{L}_\textit{SSL} (z_i, z_\textit{avg}) \ ,
\end{equation}
where $z_\textit{avg} = \frac{\sum_{i=1}^N z_i}{N}$ is the average of the patch representations, $\alpha$ and $\beta$ are scalar hyperparameters, and $\mathcal{L}_\textit{TCR}$ is Total Coding Rate loss, defined as: 
\begin{equation}
    \mathcal{L}_\textit{TCR}([z_1, ..., z_N] = Z) = \frac{1}{2} \log \det \left(\textit{I} + \frac{d}{b\epsilon^2}ZZ^\top \right) \ ,
\end{equation}
where $b$ is the batch size, $\epsilon>0$ a chosen size of distortion, and $d$ the dimension of the feature vectors. Compared to existing multi-patch strategies, our formulation acts as a plug-in for other SSL models, thus being able to exploit and improve over the advantages they already provide.

We now describe the application of CMP to two popular SSL methods: SimSiam \cite{chen2020simsiam} and BYOL \cite{grill2020byol}.

\paragraph{SimSiam-CMP.}
SimSiam uses an additional projector network, called the \textit{predictor} $P$, to further project the two representations $z_1, z_2$. Given the cosine similarity function $S_c$, SimSiam loss reads:
\begin{equation}
    \mathcal{L}_\textit{SimSiam} = -S_c(\texttt{stopgradient}(z_1), p_2) -S_c(\texttt{stopgradient}(z_2), p_1) \ ,
\end{equation}
where representation collapse is avoided by preventing gradient flow through $z_1$ and $z_2$. We design SimSiam-CMP, which applies our CMP on top of SimSiam, with the following loss:
\begin{equation}
    \mathcal{L}_\textit{SimSiam-CMP} = \beta\mathcal{L}_{TCR}([z_1, ..., z_N]) + \alpha \sum_{i=1}^N -S_c(\texttt{stopgradient}(z_\textit{avg}), p_i) \ .
\end{equation}

\paragraph{BYOL-CMP.} Like SimSiam, BYOL also uses the predictor $P$. BYOL keeps a copy of the encoder $\theta$, called $\theta'$, updated via the Exponential Moving Average (EMA) of $\theta$ weights. Given $z_1', z_2'$ the representations extracted with $\theta'$, BYOL loss is defined as follows:
\begin{equation}
    \mathcal{L}_\textit{BYOL} = \text{MSE}(\overline{z}_1', \overline{p}_2) + \text{MSE}(\overline{z}_2', \overline{p}_1) \ ,
\end{equation}
where MSE is the mean squared error and $\overline{z}_1'$, $\overline{z}_2'$, $\overline{p}_1$, $\overline{p}_2$ are the $\ell_2$-normalized $z_1'$, $z_2'$, $p_1$, $p_2$, respectively.
We adopt the same approach as for SimSiam and propose BYOL-CMP with the following loss:
\begin{equation}
        \mathcal{L}_\textit{BYOL-CMP} = \beta \mathcal{L}_{TCR}([z_1, ..., z_N]) + \alpha \sum_{i=1}^N \text{MSE}(\overline{z}_\textit{avg}', \overline{p}_i) \ .
\end{equation}
Unlike SimSiam, BYOL-CMP leverages the normalized features encoded by $\theta'$ instead of the ones encoded by $\theta$.

\section{Experiments}
\begin{table}[t]
    \centering
    \small
    \begin{tabular}{ccccc}
    \toprule
        \multirow{2}{*}{\textsc{SSL Method}}& \multirow{2}{*}{\textsc{Strategy}} & \multirow{2}{*}{$\mathcal{M}$ \textsc{size}} & \multicolumn{2}{c}{\textsc{Probing Accuracy}} \\
         &  & & CIFAR-100 & ImageNet100 \\
        
        \cmidrule(rl){1-1}\cmidrule(rl){2-2} \cmidrule(rl){3-3} \cmidrule(rl){4-5}
        EMP-SSL & - & $0$ & $28.5 \pm {\scriptstyle 0.6}$ & $32.7 \pm {\scriptstyle 1.1}$ \\
        \midrule
        \multirow{5}{*}{SimSiam}
        & finetuning & $0$ & $17.9 \pm {\scriptstyle 0.7}$ & $11.7 \pm {\scriptstyle 0.5}$ \\
        & Reservoir ER & $500$ & $29.1 \pm {\scriptstyle 0.2}$ & $33.5 \pm {\scriptstyle 0.5}$ \\
        & Reservoir ER & $2000$ & $27.6 \pm {\scriptstyle 0.4}$ & $\mathbf{39.5 \pm {\scriptstyle 0.5}}$ \\
        & FIFO ER & $90$  & $25.5 \pm {\scriptstyle 0.5}$ & $30.0 \pm {\scriptstyle 1.7}$ \\
        & \textbf{CMP} (our) & $0$ & $\mathbf{30.2 \pm {\scriptstyle 0.7}}$ & $33.3 \pm {\scriptstyle 0.7}$\\
        \midrule
        \multirow{5}{*}{BYOL}
        & finetuning & $0$ & $13.3 \pm {\scriptstyle 0.0}$ & $11.3 \pm {\scriptstyle 0.2}$ \\
        & Reservoir ER & $500$ & $34.0 \pm {\scriptstyle 0.5}$ & $33.9 \pm {\scriptstyle 0.2}$ \\
        & Reservoir ER &  $2000$&  $32.0 \pm {\scriptstyle 0.1}$ & $40.3 \pm {\scriptstyle 0.7}$ \\
        & FIFO ER  & $90$ & $27.6 \pm {\scriptstyle 0.7}$ & $29.8 \pm {\scriptstyle 0.8}$ \\
        & \textbf{CMP} (our) & $0$ & $\mathbf{34.6 \pm {\scriptstyle 0.7}}$ & $\mathbf{46.3 \pm {\scriptstyle 0.3}}$ \\
    \bottomrule
\end{tabular}
    \caption{Linear probing accuracy on Split CIFAR-100 and Split ImageNet100. We report results mean and standard deviation across 3 runs. Best in \textbf{bold}.}\label{tab:results-b200}
\end{table}

% \begin{table}[b]
%     \label{tab:my_label}
%     \caption{Comparing encoders trained on an OCSSL stream vs enoders trained on a supervised OCL stream, by executing linear probing on all.}
%     \centering
%     \begin{tabular}{cc}
%     \toprule
%        \textsc{Strategy} & \textsc{Accuracy} \\
%         \midrule
%         Supervised w/ ER & \\
%         Others? & \\

%         \midrule
%         EMP & \\
%         \textbf{SimSiam MultiPatch} & \\
%         \textbf{BYOL MultiPatch} & \\
%     \bottomrule
%     \end{tabular}
% \end{table}

We conducted experiments on two OCSSL class-incremental streams: Split CIFAR-100 and Split ImageNet100, with 20 splits each. We set the streaming minibatch size $b_s$ to 10 (i.e., number of examples available at each training iteration), and, for each, CMP extracts 20 patches, resulting in a final batch size of $200$.
We chose ResNet-18 as the backbone network, optimized using SGD with $0.9$ momentum and $1 \times 10^{-4}$ weight decay. We conducted a grid search to select learning rate,  $\alpha$ and $\beta$ on a held-out set of 10\% validation data.
After training on the data stream, evaluation was performed via linear probing, as commonly done in SSL. The probe was trained with a $0.05$ learning rate, reduced by one third whenever the validation accuracy stopped improving. Training stopped after 100 epochs or when the learning rate decreased below $1\mathrm{e}{-4}$.
% For CMP and EMP-SSL we applied the average bag-of-feature approach for evaluation, similarly to \cite{tong2023emp, chen2022bagssl}.

\paragraph{Baselines.} We compared both SimSiam-CMP and BYOL-CMP against their base SSL variants. We also pair them with ER strategies that extend the streaming batch size with $90$ replay samples at each step, resulting in the same final batch size of $200$ as CMP, after the standard SSL two-views augmentations.
ER buffers are either filled with reservoir sampling (memory size 2000 and 500) or with a FIFO buffer, which is composed of the last $90$ streaming examples (updated with FIFO policy). In FIFO ER, at each iteration, the model is trained on the buffer plus the current streaming batch. This allows to keep only the most recent examples into consideration, thus comparing CMP with an ER strategy with minimal past bias.

\paragraph{Results.} In Tab. \ref{tab:results-b200} we report the linear probing accuracy computed at the end of training on all the stream. CMP surpasses all ER-based methods, except for SimSiam with $\mathcal{M}$ size = 2000. This is surprising, as ER is often considered as the best-performing method in CL, if not even a requirement needed to mitigate the small amount of streaming examples available at each step. In particular, BYOL-CMP is the best-performing method across both benchmarks. This clarifies and explains the advantage of CMP compared to EMP, since CMP is able to combine the benefits of the EMA-updated encoder from BYOL with the multiple views per minibatch of CMP. A similar protocol could improve the performance of other SSL methods as well.\\
Our hypothesis that increasing the minibatch size at each training step is crucial for the final performance is validated by CMP, which achieves more than double the accuracy of simple fine-tuning (where the minibatch is not extended). constraints on buffer availability hinder performance, confirming that CMP enhances fast adaptation in OCSSL scenarios.\\
Interestingly, in CIFAR-100 having a larger ER buffer (2000 vs. 500) reduces the performance, while it is not the case for ImageNet100. Our hypothesis is that when a smaller set of samples is used for replay, the model is able to converge quicker than with a larger number (as each single example is used in more training iterations). This assumptions holds for simpler datasets, like CIFAR-100, which need less generalization abilities and for which few samples are representative enough. This is not true in more complex benchmarks, such as ImageNet100, which require more generalization capabilities and thus benefit from a more diverse set of examples.\\
Overall, we find CMP to be an effective solution for OCSSL scenarios, offering a competitive replay-free approach that is competitive or superior with respect to existing replay-based OCSSL approaches, even the ones also leveraging multiple patches like EMP-SSL.

\section{Conclusion}
We presented CMP, a self-supervised method designed for OCL streams, where no information about data drift is available and where only a few examples can be accessed at a time. Our results show that CMP is not only able to learn effective representations online, but it also surpasses the performance of replay strategies, which are commonly considered an essential component of OCL. Instead, our CMP achieves a strong performance without any external buffer, hinting at the fact that once an SSL method is able to learn online, it is also able to mitigate forgetting without the need for revisiting previous samples. 
% Overall, contrary to previous works, our study proves that fast adaptation in an OCL context is possible even in the absence of a replay buffer.

% Overall, our study on OCL suggests that fast convergence and quick adaptation are stronger desiderata than memory consolidation and mitigation of forgetting when learning from non-stationary online data streams. 

% ****************************************************************************
% BIBLIOGRAPHY AREA
% ****************************************************************************

\begin{footnotesize}

% IF YOU DO NOT USE BIBTEX, USE THE FOLLOWING SAMPLE SCHEME FOR THE REFERENCES
% ----------------------------------------------------------------------------

% ----------------------------------------------------------------------------

% IF YOU USE BIBTEX,
% - DELETE THE TEXT BETWEEN THE TWO ABOVE DASHED LINES
% - UNCOMMENT THE NEXT TWO LINES AND REPLACE 'Name_Of_Your_BibFile'

% \bibliographystyle{unsrt}
\bibliographystyle{abbrv}
\bibliography{mybib}

\begin{thebibliography}{10}

\bibitem{bardes2022vicreg}
A.~Bardes, J.~Ponce, and Y.~LeCun.
\newblock Vicreg: Variance-invariance-covariance regularization for self-supervised learning, 2022.

\bibitem{caron2021swav}
M.~Caron, I.~Misra, J.~Mairal, P.~Goyal, P.~Bojanowski, and A.~Joulin.
\newblock Unsupervised learning of visual features by contrasting cluster assignments, 2021.

\bibitem{chaudhry2019er_online}
A.~Chaudhry, M.~Rohrbach, M.~Elhoseiny, T.~Ajanthan, P.~K. Dokania, P.~H.~S. Torr, and M.~Ranzato.
\newblock On tiny episodic memories in continual learning, 2019.

\bibitem{chen2020simclr}
T.~Chen, S.~Kornblith, M.~Norouzi, and G.~Hinton.
\newblock A simple framework for contrastive learning of visual representations, 2020.

\bibitem{chen2020simsiam}
X.~Chen and K.~He.
\newblock Exploring simple siamese representation learning, 2020.

\bibitem{chen2022bagssl}
Y.~Chen, A.~Bardes, Z.~Li, and Y.~LeCun.
\newblock Bag of image patch embedding behind the success of self-supervised learning.
\newblock {\em arXiv preprint arXiv:2206.08954}, 2022.

\bibitem{cossu2022pretraining}
A.~Cossu, T.~Tuytelaars, A.~Carta, L.~Passaro, V.~Lomonaco, and D.~Bacciu.
\newblock Continual pre-training mitigates forgetting in language and vision, 2022.

\bibitem{fini2022cassle}
E.~Fini, V.~G.~T. da~Costa, X.~Alameda-Pineda, E.~Ricci, K.~Alahari, and J.~Mairal.
\newblock Self-supervised models are continual learners, 2022.

\bibitem{gomezvilla2022pfr}
A.~Gomez-Villa, B.~Twardowski, L.~Yu, A.~D. Bagdanov, and J.~van~de Weijer.
\newblock Continually learning self-supervised representations with projected functional regularization, 2022.

\bibitem{grill2020byol}
J.-B. Grill, F.~Strub, F.~Altch\'e, C.~Tallec, P.~H. Richemond, E.~Buchatskaya, C.~Doersch, B.~A. Pires, Z.~D. Guo, M.~G. Azar, B.~Piot, K.~Kavukcuoglu, R.~Munos, and M.~Valko.
\newblock Bootstrap your own latent: A new approach to self-supervised learning, 2020.

\bibitem{gui2023sslsurvey}
J.~Gui, T.~Chen, J.~Zhang, Q.~Cao, Z.~Sun, H.~Luo, and D.~Tao.
\newblock A survey on self-supervised learning: Algorithms, applications, and future trends, 2023.

\bibitem{mai2021onlinesurv}
Z.~Mai, R.~Li, J.~Jeong, D.~Quispe, H.~Kim, and S.~Sanner.
\newblock Online continual learning in image classification: An empirical survey, 2021.

\bibitem{purushwalkam2022minred}
S.~Purushwalkam, P.~Morgado, and A.~Gupta.
\newblock The challenges of continuous self-supervised learning.
\newblock In S.~Avidan, G.~Brostow, M.~Ciss{\'e}, G.~M. Farinella, and T.~Hassner, editors, {\em Computer Vision -- ECCV 2022}, pages 702--721, Cham, 2022. Springer Nature Switzerland.

\bibitem{rebuffi2017icarl}
S.-A. Rebuffi, A.~Kolesnikov, G.~Sperl, and C.~H. Lampert.
\newblock icarl: Incremental classifier and representation learning.
\newblock In {\em Proceedings of the IEEE conference on Computer Vision and Pattern Recognition}, pages 2001--2010, 2017.

\bibitem{soutifcormerais2023comprehensive}
A.~Soutif-Cormerais, A.~Carta, A.~Cossu, J.~Hurtado, H.~Hemati, V.~Lomonaco, and J.~V. de~Weijer.
\newblock A comprehensive empirical evaluation on online continual learning, 2023.

\bibitem{tong2023emp}
S.~Tong, Y.~Chen, Y.~Ma, and Y.~Lecun.
\newblock Emp-ssl: Towards self-supervised learning in one training epoch.
\newblock {\em arXiv preprint arXiv:2304.03977}, 2023.

\bibitem{yu2023scale}
X.~Yu, Y.~Guo, S.~Gao, and T.~Rosing.
\newblock Scale: Online self-supervised lifelong learning without prior knowledge, 2023.

\end{thebibliography}

\end{footnotesize}

% ****************************************************************************
% END OF BIBLIOGRAPHY AREA
% ****************************************************************************

\end{document}